\documentclass[runningheads]{llncs}
\usepackage{graphicx}
\begin{document}
\title{An Interpretable Machine Vision Approach to Human Activity Recognition using Photoplethysmograph Sensor Data}
\titlerunning{Machine Vision for Human Activity Recognition}
%
\author{Eoin Brophy\orcidID{0000-0002-6486-5746} \and
Jos\'e Juan Dominguez Veiga\orcidID{0000-0002-6634-9606} \and
Zhengwei Wang\orcidID{0000-0001-7706-553X} \and
Alan F. Smeaton\orcidID{0000-0003-1028-8389} \and
Tom\'as E. Ward\orcidID{0000-0002-6173-6607}} 
\authorrunning{Brophy et al.}
%
\institute{Insight Centre for Data Analytics, \newline
Dublin City University, Glasnevin, Dublin, Ireland}
\maketitle              
\pagenumbering{gobble} 

\begin{abstract}
The current gold standard for human activity recognition (HAR) is based on the use of cameras. However, the poor scalability of camera systems renders them impractical in pursuit of the goal of wider adoption of HAR in mobile computing contexts. Consequently, researchers instead rely on wearable sensors and in particular inertial sensors. A particularly prevalent wearable is the smart watch which due to its integrated inertial and optical sensing capabilities holds great potential for realising better HAR in a non-obtrusive way. This paper seeks to simplify the wearable approach to HAR through determining if the wrist-mounted optical sensor alone typically found in a smartwatch or similar device can be used as a useful source of data for activity recognition. The approach has the potential to eliminate the need for the inertial sensing element which would in turn reduce the cost of and complexity of smartwatches and fitness trackers. This could potentially commoditise the hardware requirements for HAR while retaining the functionality of both heart rate monitoring and activity capture all from a single optical sensor. Our approach relies on the adoption of machine vision for activity recognition based on suitably scaled plots of the optical signals. We take this approach so as to produce classifications that are easily explainable and interpretable by non-technical users. More specifically, images of photoplethysmography signal time series are used to retrain the penultimate layer of a convolutional neural network which has initially been trained on the ImageNet database. We then use the 2048 dimensional features from the penultimate layer as input to a support vector machine. Results from the experiment yielded an average classification accuracy of 92.3\%. This result outperforms that of an optical and inertial sensor combined (78\%) and illustrates the capability of HAR systems using standalone optical sensing elements which also allows for both HAR and heart rate monitoring. Finally, we demonstrate through the use of tools from research in explainable AI how this machine vision approach lends itself to more interpretable machine learning output.

\end{abstract}
\section{Introduction}
Due to the ubiquitous nature of inertial and physiological sensors in phones and fitness trackers, human activity recognition (HAR) studies have become more widespread \cite{1}. The benefits of HAR include rehabilitation for recovering patients, activity monitoring of the elderly and vulnerable people, and advancements in human-centric applications \cite{2}.

Photoplethysmography (PPG) is an optical technique used to measure volume changes of blood in the microvascular tissue. PPG is capable of measuring heart rate by detecting the amount of light reflected/absorbed in red blood cells as this varies with the cardiac cycle. Reflected light is read by an ambient light sensor which then has its output conditioned, so a pulse rate can be determined. The pulse rate is obtained from analysis of the small alternating component (which arises from the pulsatile nature of blood flow) superimposed on the larger base signal caused by the constant absorption of light.

For usability reasons the wrist is a common site for wearables used in health and fitness contexts \cite{3}. Most smartwatches are equipped with a PPG sensor capable of measuring pulse rate. Difficulties arising in obtaining a robust physiologically useful signal via PPG can be caused by motion artefacts due to the changes in optical path length associated with disturbance of the source-detector configuration. This disturbance is introduced by haemodynamic effects and gross motor movements of the limbs \cite{4}. This can lead to an incorrect estimation of the pulse rate. Reduction in motion artefacts can be achieved using a range of techniques, for example, through the choice of physical attachment technique or through adaptive signal filtering methods based on an estimation of the artefact source from accelerometer-derived signals \cite{5}.

In this study we sought to exploit the motion artefact to infer human activity type from the PPG signals collected at the wrist. Our hypothesis was that there is sufficient information in the disturbance induced in the source-detector path to distinguish different activities through the use of a machine learning approach. In recent years, capabilities of machine learning methods in the field of image recognition has increased dramatically \cite{6}. Building on these advancements in image recognition would allow for simplification of wearables involved in HAR. We chose an image-based approach to the machine learning challenge as this work is part of a larger scope effort to develop easily deployed AI (artificial intelligence) which can be used and interpreted by end users who do not have signal processing expertise.

This paper extends previous work completed in \cite{7}. The additional contributions of this paper are improved classification through the use of a hybrid classifier approach and more significantly to demonstrate through the use of tools from research in explainable artificial intelligence (XAI) how this machine vision approach lends itself to more interpretable machine learning output.

\section{Related Work}
Convolutional neural networks (CNNs) have been used since the 1990s and were designed to mimic how the visual cortex of the human brain processes and recognises images \cite{8}. CNNs extract salient features from images at various layers of the network. They allow implementation of high-accuracy image classifiers given the correct training without the need for in-depth feature extraction knowledge. 

The current state of the art in activity recognition is based on the use of cameras. Cameras allow direct and easy capture of motion but this output requires significant processing for recognition of specific activities. Inertial sensing is another popular method used in HAR. To achieve the high accuracy of the inertial sensing systems shown in \cite{3}, a system consisting of multiple sensors is required, compromising functionality and scalability. The associated signal processing is not trivial and singular value decomposition (SVD), truncated Karhunen-Lo\`eve transform (KLT), Random Forest (RF) and Support Vector Machines (SVM) are examples of feature extraction and machine learning methods that have been applied to HAR.

Inertial sensor data paired with PPG are amongst the most suitable sensors for activity monitoring as they offer effective tracking of movement actions as well as relevant physiological parameters such as heart rate. They also have the benefit of being easy to deploy. Mehrang et al. used RF and SVM machine learning methods for a HAR classifier on combined accelerometer and PPG data achieving an average recognition accuracy of 89.2\% and 85.6\% respectively \cite{9}.

The average classification accuracy of the leading modern feature extraction and machine learning methods for singular or multiple inertial sensors range from 80\% to 99\% \cite{3}. However, this can require up to 5 inertial measurement units located at various positions on the body.

Even with the success of deep learning, similar criticisms that plagued past work on neural networks, mainly in terms of being uninterpretable, are recurring. There is no unanimous definition of interpretability, but works like Doshi-Velez and Kim \cite{doshi} try to define the term and provide a way of measuring how interpretable a system is. Issues such as algorithmic fairness, accountability and transparency are a source of current research. An example for the need of such work is the European Union’s “right to explanation” \cite{goodman} in which users can demand to be explained how an algorithmic decision that affects them has been made. The ramifications of these implicitly present issues with governance, policy and privacy demonstrate a need for regulation of explainability in the near future. In this paper the authors use the term interpretability in terms of transparency and accountability.

\section{Methodology}
\subsection{Data Collection} 
We use a dataset collected by Deleram Jarchi and Alexander J. Casson and is freely available from Physionet \cite{10}. This dataset comprises PPG recordings taken from 8 patients (5 female, 3 male) aged between 22-32 (mean age of 26.5), during controlled exercises on a treadmill and an exercise bike. Data was recorded using a wrist-worn PPG sensor attached to the Shimmer 3 GSR+ unit for an average period of 4-6 minutes with a maximum duration of 10 minutes. A frequency of 256 Hz was used to sample the physiological signal. Each individual was allowed to set the intensity of their exercises as they saw fit and every exercise began from rest. The four exercises were broken down into walk on a treadmill, run on a treadmill, low and high resistance exercise bike. For the walk and run exercises the raw PPG signals required no filtering other than what the Shimmer unit provides. The cycling recordings were low-pass filtered using Matlab with a 15 Hz cut-off frequency to remove any high-frequency noise.

\subsection{Data Preparation}
The PPG data signal was downloaded using the PhysioBank ATM and plotted in Python. Signals were segmented into smaller time series windows of 8-second intervals chosen to match the time windows used in \cite{11}, which acts as a benchmark for this study. A rectangular windowing function was used to step through the data every 2 seconds, also conducted in \cite{11}. It is worth re-emphasising that a machine vision approach is being taken here, the input data to the classifier is not time series vectors but images. These images correspond to simple plots of the 8-second window produced in Python which plots and saves figures and removes all axis labels, legends and grid ticks (removing non-salient features), saving each figure as a 299x299 JPEG file. A total of 3321 images were created, of which 80\% (2657) were used for retraining, 10\% (332) for validation and 10\% (332) for testing.

\begin{figure}
\centering
\includegraphics[width=0.5\textwidth]{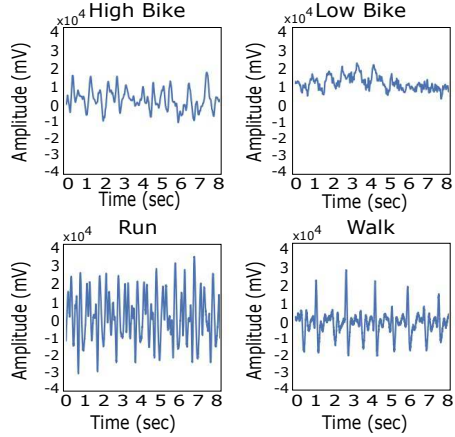}
\caption{Sample of PPG image for each activity} \label{fig1}
\end{figure}

The image files were stored in a directory hierarchy based on the movement carried out. Four sub-directories of possible classifiers were created; run, walk, high resistance bike and low resistance bike and contained within each were the images of the plotted PPG signal. In Fig~\ref{fig1} an example of a plot for each activity can be seen.

\subsection{The Network Infrastructure}
On completion of data preparation, the CNN could then be retrained. Building a neural network from the ground up is not a trivial task, requiring a multilayer implementation for even a simple perceptron \cite{8} which needs optimization of tens of thousands of parameters for even a trivial task such as handwritten digit classification \cite{12}. Also very large amounts of data is required for training. Here we avoided both problems through use of Inception-v3 which can be implemented with the TensorFlow framework and transfer learning.

TensorFlow \cite{13}, a deep learning framework, was used for transfer learning which is the concept of using a pretrained CNN and retraining the penultimate layer that does classification before the output. This type of learning is ideal for this study due to our relatively small dataset \cite{14}. The results of the retraining process can be viewed using the suite of visualisation tools on TensorBoard.

Recognition of specific objects from millions of images requires a model with a large learning capacity. CNNs are particularly suitable for image classification because convolution leverages three important properties that can help improve a machine learning system: sparse interaction, parameter sharing and equivariant representations \cite{15}. These properties enable CNNs to detect small, meaningful features from input images and reduce the storage requirements of the model compared to traditional densely connected neural networks. CNNs are tuneable in their depth and breadth meaning they can have much simpler architecture, making them easier to train compared to other feedforward neural networks \cite{6}.

\subsection{Retraining and Using the Network}
A pretrained CNN, Google's Inception-v3, was used which has been trained on ImageNet, a database of over 14 million images and will soon be trained on up to 50 million images [16]. To retrain a network, we use the approach taken by Dominguez Veiga et. al \cite{17}. The experiment was completed again using the features from the retrained Inception-v3 penultimate layer. The 2048 dimensional features were extracted from this layer and used as input into a radial basis function SVM classifier as in \cite{9}, a nested cross-validation strategy was applied here to prevent biasing the model evaluations. 

The retraining process can be fine-tuned through hyperparameters which allows for optimisation of the training outcome. For the transfer learning approach the default parameters were used except for the number of training steps which were changed from the default of 4,000 steps to 10,000 steps. Selection of this number of iterations allowed for the loss function (cross-entropy) to sufficiently converge, avoiding over-fitting.

\section{Results}
\begin{figure}
\includegraphics[width=\textwidth]{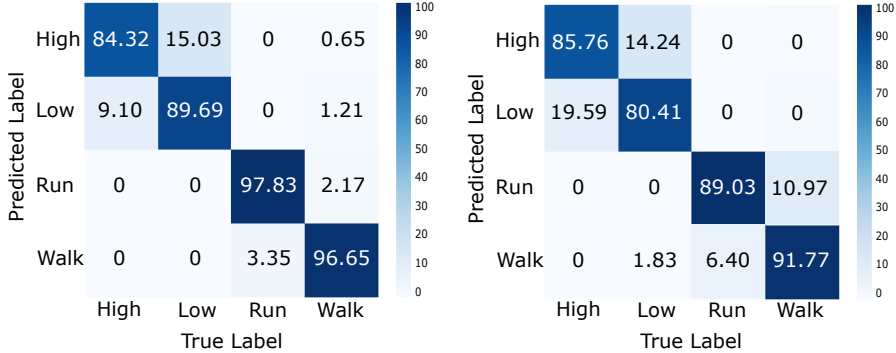}
\caption{Confusion matrix for CNN-SVM and for transfer learning approach} \label{fig2}
\end{figure}

The average classification accuracy of the transfer learning approach was shown to be 88.3\% and as can be seen in Fig~\ref{fig2}(b) the confusion matrix, demonstrates the accuracy for correctly classifying the test set of PPG images.

The CNN-SVM achieved an average classification accuracy of 92.3\%, see Fig~\ref{fig2}(a) for the confusion matrix associated with this method. The confusion matrices for these deep learning approaches graphically demonstrates some of the issues classifying the low-resistance bike exercise, where it was misclassified as high-resistance 19.59\% of the time. An increase 4.8\% classification accuracy was achieved using the CNN-SVM approach with nested cross-validation over the transfer learning approach.

Fig~\ref{fig4} shows two misclassified examples of each class. Based on the plots shown in Fig~\ref{fig4}, the errors may have been from a loose wrist strap or excessive movement of the arms. In some circumstances, the signal can be seen to be cut off which indicates gross movement of the limbs for those time instances \cite{4}.

\begin{figure}[t]
\centering
\includegraphics[width=0.5\textwidth]{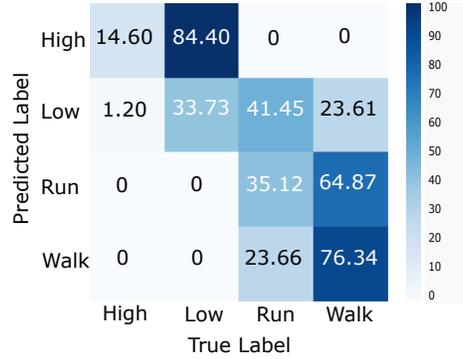}
\caption{Confusion matrix for feature extraction approach} \label{fig3}
\end{figure}

\begin{figure}
\centering
\includegraphics[height=4.5cm, width=0.85\textwidth]{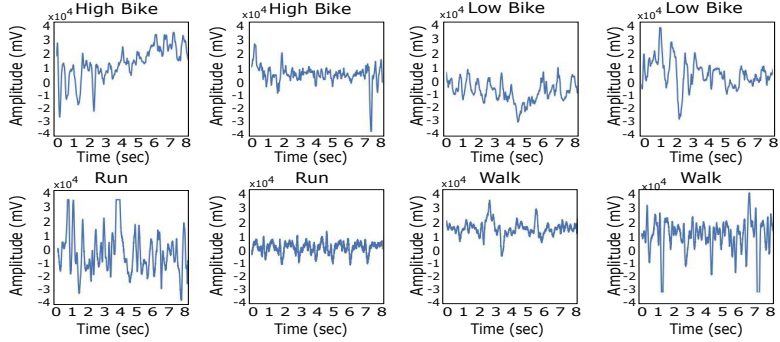}
\caption{Sample of eight misclassified images} \label{fig4}
\end{figure}

A study conducted by Biagetti et al. used feature extraction and reduction with Bayesian classification on time series signals \cite{11}. Their technique focused on using singular value decomposition and truncated Karhunen-Lo\`eve transform. Their study used the same time series dataset and was designed to present an efficient technique for HAR using PPG and accelerometer data. Below, in Fig~\ref{fig3} the confusion matrix for their results can be seen, using just the PPG signal for feature extraction and classification.

The feature extraction approach for determining HAR using just the PPG yields an overall accuracy of 44.7\%. This shows a reduced classification performance versus the deep learning method employed in this paper. Although our CNN-SVM achieved greater accuracy of over 47 percentage points (92.3\% vs. 44.7\%), Biagetti et al. in the same paper combined the PPG and accelerometer data to bring their classifier accuracy rate to 78\%. We are able to produce very competitive accuracy without the use of an accelerometer, i.e. through the optical signal only.

\section{Discussion}
Applying the CNN-SVM method to the PPG dataset leads to an accuracy of 92.3\%. This result outperforms the combined PPG and accelerometer data for HAR using SVD and KVL (92.3\% vs. 78\%) and of course is much more accurate than the PPG only result (92.3\% vs. 44.7\%) \cite{11}. This is a highly competitive result and suggests that simpler wearables based on optical measurements only could yield much of the functionality achievable with more sophisticated, existing multi-modal devices. Of course, the addition of an inertial sensor will always produce more information and therefore more nuanced activity recognition. However, for the types of activity recognition commonly sought in clinical, health and fitness applications a surprisingly good performance can be extracted from a very simple optical measurement.

\subsection{Limitiations}
Better understanding of the hyperparameters of the transfer learning prior to the CNN-SVM may lead to higher average classification accuracy than the one achieved in this paper (92.3\%). The results generated in this paper are based on data from a small sample size which contained a low number of individuals and this may affect the classifier if implemented of new, previously unseen data.

\subsection{Visualising the Network}
Due to the increase in the use of machine learning and neural networks in the last number of years there has been a growing demand to make their working more transparent and understandable by end users. Often, neural network models are opaque and difficult to visualise. A lack of insight into the model results in diminished trust on the part of the end user. People want to know why and when the model succeeds and how to correct errors \cite{19}.

CNNs can appear to operate as a black box. In our case, the input are images, the layers process this and intermediate layers to do their work; extracting patterns, edges and more abstract features. Then a decision is output to the user foregoing any explanation as to why the particular decision was reached e.g. ``This image is a dog". Consequently there is emerging and growing efforts to make such AI systems more explainable. For example, the US Defense Advanced Research Projects Agency (DARPA) are currently working on an explainable model with an explainable interface, i.e. ``This is a dog because of the snout, ears, body shape etc." This explainable interface can help end-users better accept the decisions made by the AI system.

A number of works have been able to produce high quality feature visualisation techniques, with the goal of making CNNs more explainable. The work done in \cite{20} has the capability to demonstrate the function of the layers within a CNN, how it builds up extracting edges, textures and patterns to extracting parts of objects. The work done by Olah, et al., shows us what a CNN `sees' and how it builds up its layers as we go deeper into the network.

However, end users are still unable to intuitively determine why the input caused the output. The information retaining to what feature in the input image is a significant discriminating factor in the classification decision is unavailable. The issue here is to be able to explain the decision making process rather than visually representing what each layer extracts. A simple, visual representation of the decision can be a very effective method of explainable artificial intelligence.

Class activation maps (CAM) are one such visual method capable of aiding in the explanation of the decision making process. CAMs indicate particularly discriminate regions of an image that are used by a trained CNN to determine the prediction of the image for a given layer \cite{21}, in other words what region in the input image is relevant to the classification. The CAM for this paper uses the global average pooling layer of the retrained Inception-v3 model.

\begin{figure}
\centering
\includegraphics[height=8.5cm,width=8.5cm]{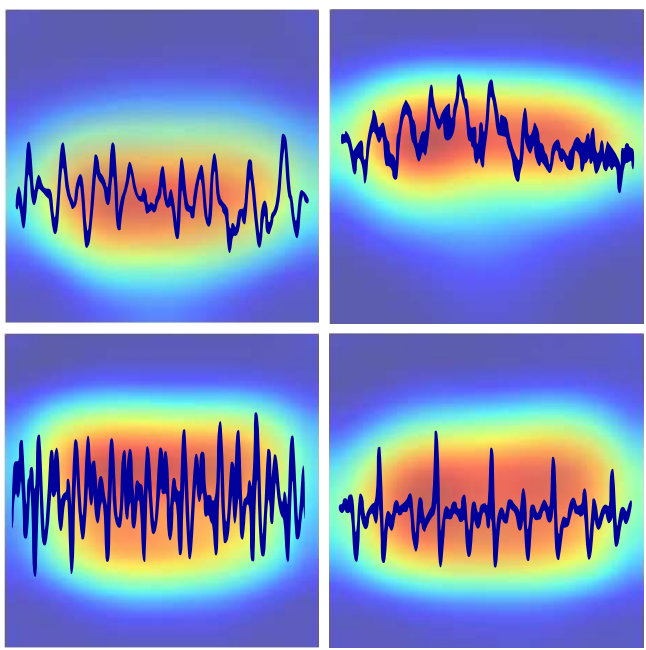}
\caption{Average CAMs for High, Low, Run and Walk classes} \label{fig5}
\end{figure}

Fig~\ref{fig5} shows the CAM for the retrained network on all of the input classes; high, low, run and walk respectively, they are representative of the average CAMs for their class. Red areas indicate high activations while the dark blue areas indicate little to no activation. Activations across all four classes demonstrate that the model is correctly discriminating the plotted signal from the background of the image.  The location of the activations can be used to indicate the inter-class variations. Each of the classes produces a distinct CAM from one another, the difference in these activations can be used to indicate how a class was classified.  For a human observer we can say the class is such due to it's amplitude or frequency. The inter-class variation in the CAM across these plots may provide an explainable interface into the decision making process taken by the CNN regarding the image classification problem. ``The activity was classified as a walk due to the location of the content activations not present in other classes."

\begin{figure}
\centering
\includegraphics[height=6.5cm,width=6.5cm]{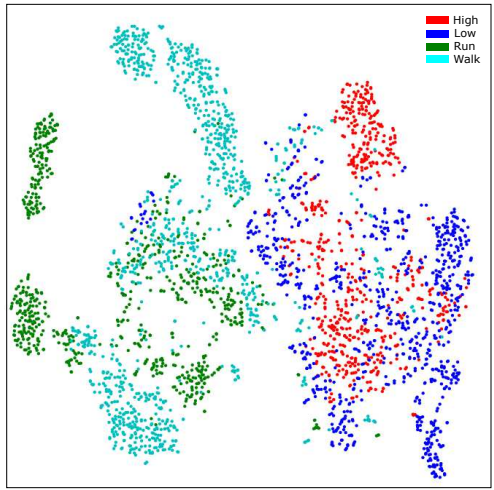}
\caption{t-SNE visualisations of CNN codes} \label{fig6}
\end{figure}

Another XAI approach that can be adopted is the use of t-Distributed Stochastic Neighbor Embedding (t-SNE) \cite{22}. This dimensionality reduction technique is well suited to the visualisation of high-dimensional datasets such as the 2048 dimensional feature vector describing an image in the penultimate layer of Inception-v3. Using t-SNE on Inception-v3 global average pooling layer allows visualisation of the dimensionally reduced feature vectors for all classes. Fig~\ref{fig6} demonstrates a clustering effect of the data, close groupings of the same class demonstrate the similarity between features available in that class. This allows a user to see the where features for one class appears in relation to the other classes in the trained model. t-SNE produces visualisations that are interpretable by the user in relating the output to the input. One way to help the user comprehend the visualisation of this technique is to place a decision boundary over the output. A feature found within the decision boundary for one class in particular would explain why the input generates the output. Conversely, it might raise further questions if the feature appears outside the boundary. It is also possible to graph one feature against another for a simple dataset (sepal vs petal length \cite{23}). However, this has not been investigated for this dataset yet and this will form part of our future work.

\section{Conclusion}
The application of transfer learning in computer vision for human activity recognition is a novel approach to extracting new useful information from wrist-worn PPG sensors which are conventionally used for heart rate monitoring. We show how CNN suitably configured can create powerful classifiers for HAR applications based on simple images of PPG time series plots \cite{6}.

A great benefit of the deep learning approach adopted here is its performance and relative simplicity. Users of this system need not possess a signal processing background to understand the approach and this opens up the possibility that non-experts can develop their own HAR classification applications more readily. Pathways for HAR using deep learning are beginning to be explored on a larger scale thanks to the simplicity of the transfer learning approach, significantly reducing the development time of a suitable CNN. Furthermore this new approach to HAR classification will allow for the easier testing of hypotheses relating to HAR with wearable sensors. The development of AI applications and platforms in the last number of years have led to significant progressions particularly in the domain of image classification problems. These black box AI techniques have a need for more explainable methods to reinforce the machines decision making process. The activation maps (Fig~\ref{fig5}) in conjunction with the t-SNE features (Fig~\ref{fig6}) have the potential to help with this explainable artificial intelligence issue. Inter-class variations in activations have the capability to provide an explainable interface to the end user; ``The exercise was classified as a run because of the high amplitude, high frequency components present in your signal". 

The presented process allows activity classification models to be constructed using PPG sensors only, potentially eliminating the need for an inertial sensor set and simplifying the overall design of wearable devices.

\subsubsection{Acknowledgements} This work was part-funded by Science Foundation Ireland
under grant number SFI/12/RC/2289 and by SAP SE.

%
%

%
%
%
%

\bibliography{references} 
\bibliographystyle{splncs04}

\end{document}